\documentclass[sigconf,screen,nonacm]{acmart}
\author{Jingkun Chen}
\affiliation{%
  \institution{Northwestern Polytechnical University}
  \country{China}
}

\author{Ruoshi Xu}
\affiliation{%
  \institution{Southern University of Science and Technology}
  \country{China}
}

\author{Mingqi Gao}
\affiliation{%
  \institution{The University of Sheffield}
  \country{United Kingdom}
}

\author{Shengda Luo}
\affiliation{%
  \institution{Hengqin Laboratory}
  \country{China}
}

\author{Jungong Han}
\affiliation{%
  \institution{Tsinghua University}
  \country{China}
}
\AtBeginDocument{%
  }

\usepackage{array}

\usepackage{booktabs}

\usepackage{tabularx}
\usepackage[table]{xcolor}
\definecolor{NatHeader}{HTML}{ECEFF3} 
\definecolor{NatRow}{HTML}{F7F7F7}    
\definecolor{NatHi}{HTML}{FFF2C6}     

\definecolor{NatBlue1}{HTML}{DCEBFF}  
\definecolor{NatBlue2}{HTML}{EAF2FF}  
\definecolor{NatGreen1}{HTML}{DCF3E6} 
\definecolor{NatGreen2}{HTML}{EAF7EF} 
\definecolor{NatPurple}{HTML}{EDE3FF} 
\definecolor{NatCyan}{HTML}{DFF5F8}   
\definecolor{NatPeach}{HTML}{FFE6D6}  
\definecolor{NatSand}{HTML}{FFF3E6}   
\definecolor{NatOurs}{HTML}{E6F6FF} %

\begin{document}

\title{Reinforcing 3D Understanding in Point-VLMs via Geometric Reward Credit Assignment}

\begin{abstract}
Point-Vision-Language Models promise to empower embodied agents with executable spatial reasoning, yet they frequently succumb to geometric hallucination where predicted 3D structures contradict the observed 2D reality. We identify a key cause of this failure not as a representation bottleneck but as a structural misalignment in reinforcement learning, where sparse geometric tokens are drowned out by noisy and broadcasted sequence-level rewards. To resolve this causal dilution, we propose Geometric Reward Credit Assignment, a framework that disentangles holistic supervision into field-specific signals and routes them exclusively to their responsible token spans. This mechanism transforms vague feedback into precise gradient updates and effectively turns generic policy optimization into targeted structural alignment. Furthermore, we internalize physical constraints via a Reprojection-Consistency term which serves as a cross-modal verifier to penalize physically impossible geometries. Validated on a calibrated benchmark derived from ShapeNetCore, our approach bridges the reliability gap by boosting 3D KPA from 0.64 to 0.93, increasing 3D bounding box intersection over union to 0.686, and raising reprojection consistency scores to 0.852. Crucially, these gains are achieved while maintaining robust 2D localization performance, marking a meaningful step from plausible textual outputs toward physically verifiable spatial predictions.
\end{abstract}





\maketitle
 
\section{Introduction}

Vision-Language Models have become strong at scene understanding, dialogue, and instruction following~\cite{wang2024qwen2,bai2025qwen3, zhu2025internvl3}. However, these abilities do not directly translate to embodied use. For a system to operate in the physical world, it is not enough to describe what is present or answer a question in natural language. The model must also recover spatial structure in a form that is metrically meaningful and grounded in the observation. This requirement is especially important for point-aware vision-language systems, where the output often contains structured geometric fields such as 2D boxes, 3D boxes, and keypoints together with text~\cite{chen2024grounded,yu2025inst3d}. In such settings, a prediction is useful only when the geometric fields and the language refer to the same physical explanation of the scene.

Point-Vision-Language Models move in this direction by combining image evidence, 3D geometry, and language generation in a single autoregressive model~\cite{chen2024ll3da,qi2024gpt4point}. This design provides a unified interface for grounded dialogue and spatial reasoning. In practice, however, these models exhibit a recurring failure mode. The model may produce a fluent answer and even a reasonable 2D localization, while the corresponding 3D prediction is not compatible with the same observation. When the predicted 3D structure is projected back to the image plane, its footprint often disagrees with the model's own 2D output. We refer to this phenomenon as geometric hallucination. The issue is not limited to small perturbations in the 3D prediction. Rather, the full prediction becomes internally inconsistent, which makes it hard to trust in downstream tasks that require physically valid grounding, a concern closely related to recent analyses of grounding reliability and hallucination in 3D-LLMs~\cite{yang20253d}.

\begin{figure}[t]
\centering
\includegraphics[width=\columnwidth]{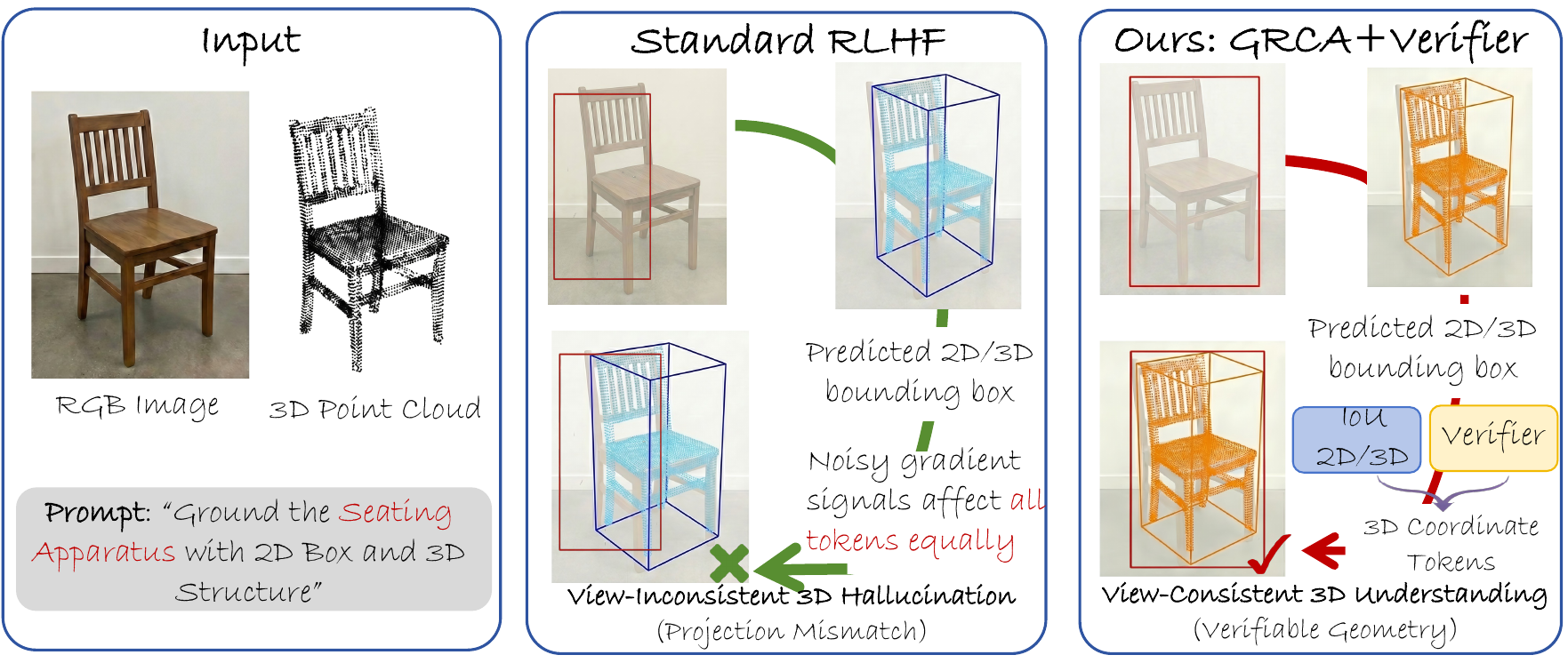}
\caption{Resolving geometric hallucination in point-based VLMs via GRCA. Compared with broadcasted-reward GRPO, GRCA with an RPC term provides targeted credit assignment to geometry tokens, improving view-consistent 3D grounding given an RGB image, a 3D point cloud, and a text prompt.}
\Description{Overview of the proposed method, comparing broadcast reward assignment with routed geometric reward credit assignment and showing improved 2D-3D consistent grounding.}
\label{fig:overview}
\end{figure}

A common explanation for this problem is that current models do not represent 3D geometry well enough. We believe this is only part of the story. The problem also lies in how the model is optimized after supervised fine-tuning. In structured generation, geometric correctness depends on a small number of coordinate-bearing tokens inside a much longer output sequence that also contains descriptive text, formatting symbols, and contextual language. Standard reinforcement learning methods usually assign a coarse sequence-level reward and apply it to all generated tokens~\cite{shao2024deepseekmath}. Such sequence-level supervision is poorly aligned with the requirements of structured spatial prediction. The tokens that actually determine whether a 3D prediction is correct receive supervision that is mixed with many unrelated tokens. In effect, the gradient signal that reaches the geometric fields becomes weak and noisy. This makes it difficult to correct small but important spatial errors, even when the overall response looks reasonable, which is consistent with recent work arguing for finer-grained reward assignment beyond sequence-level supervision~\cite{yin2025segmenting,zhou2025t}.

This observation suggests a simple principle. If a certain part of the output is responsible for a certain aspect of correctness, then the corresponding reward should be assigned to that part of the output rather than broadcast to the whole sequence. Based on this idea, we propose Geometric Reward Credit Assignment, or GRCA. GRCA treats the generated response as a structured JSON output, parses the geometric fields, evaluates each field with its own metric, and routes the resulting standardized reward only to the token span that produced that field. GRCA is applied during post-training without modifying the backbone architecture. The key change is in post-training, specifically in how reward is assigned. This yields an optimization objective that is better aligned with the structured nature of the task. It also concentrates supervision on sparse coordinate tokens while reducing interference with language-token optimization.

Field-specific supervision helps, but it still does not ensure that the full prediction is geometrically coherent. A model can improve its 2D box and its 3D box separately while the two remain inconsistent with each other. We therefore add a Reprojection-Consistency term, or RPC, as a global verifier. RPC checks whether the predicted 3D box, after projection under known camera parameters, agrees with the predicted 2D grounding from the same output. We treat this term as a global consistency reward rather than attributing it to any single local field, since it reflects a property of the prediction as a whole. GRCA and RPC therefore play different roles. GRCA improves how local geometric errors are corrected during training, while RPC encourages the final prediction to remain consistent across 2D and 3D.

We evaluate the proposed framework on a camera-calibrated benchmark derived from ShapeNetCore. The method substantially improves 3D grounding quality, boosting 3D KPA from 0.64 to 0.93, increasing 3D bounding box intersection over union to 0.686, and raising reprojection consistency to 0.852, while preserving strong 2D localization performance. These results suggest that the main obstacle to reliable 3D grounding in Point-VLMs is not representation alone, but how geometric supervision is delivered during post-training. Our main contributions are as follows.

\begin{itemize}
\item We identify geometric hallucination as a key reliability bottleneck in Point-VLMs, where predicted 3D structure is inconsistent with the model's own 2D grounding, and trace this failure to misaligned reward delivery during post-training.

\item We propose Geometric Reward Credit Assignment, a structure-aware reinforcement learning framework that aligns supervision with the internal structure of the output by routing field-specific rewards to the token spans that generate the corresponding geometric fields, together with a Reprojection-Consistency term that promotes globally consistent 2D and 3D predictions.

\item We demonstrate that the proposed framework substantially improves structured 3D grounding on a camera-calibrated benchmark derived from ShapeNetCore, boosting 3D reliability while maintaining strong 2D grounding performance.
\end{itemize}

\section{Related Work}

\subsection{3D Vision-Language Modeling and Instruction Following}
Early work on 3D vision-language modeling was largely centered on representation transfer from 2D foundation models. PointCLIP~\cite{zhang2022pointclip} and ULIP~\cite{xue2023ulip} align point cloud encoders with language-image embedding spaces, enabling zero-shot recognition and retrieval in 3D domains. OpenScene~\cite{peng2023openscene} and OpenMask3D~\cite{takmaz2023openmask3d} extend this line to dense scene understanding through open-vocabulary retrieval and segmentation. These methods greatly improve the semantic coverage of 3D perception, but their outputs remain fundamentally discriminative. They decide which concept best matches the scene or region, rather than producing a structured spatial hypothesis that must remain valid under scene geometry.

Recent work has moved toward generative 3D vision-language models that connect scene representations with large language models. Point-LLM~\cite{xu2024pointllm}, Point-Bind~\cite{guo2023point}, LL3DA~\cite{chen2024ll3da}, Chat-3D~\cite{wang2023chat}, 3D-LLM~\cite{hong20233d}, Grounded 3D-LLM~\cite{chen2024grounded}, Chat-Scene~\cite{huang2024chat}, Scene-LLM~\cite{fu2024scene}, and Inst3D-LMM~\cite{yu2025inst3d} show that 3D scenes can support open-ended dialogue, grounding, captioning, and reasoning in a unified generative interface. SceneVerse~\cite{jia2024sceneverse} further scales grounded 3D vision-language learning with large-scale scene-language supervision. Our setting is closer to this generative line, but it is also more constrained. The target output is not free-form language alone. It contains geometric fields whose correctness depends on metric accuracy, camera consistency, and whether the prediction can be executed as a valid spatial explanation of the scene. This difference changes the optimization problem itself, since a response can be linguistically plausible while still failing as a geometric prediction.

\begin{figure*}[t]
\centering
\includegraphics[width=\textwidth]{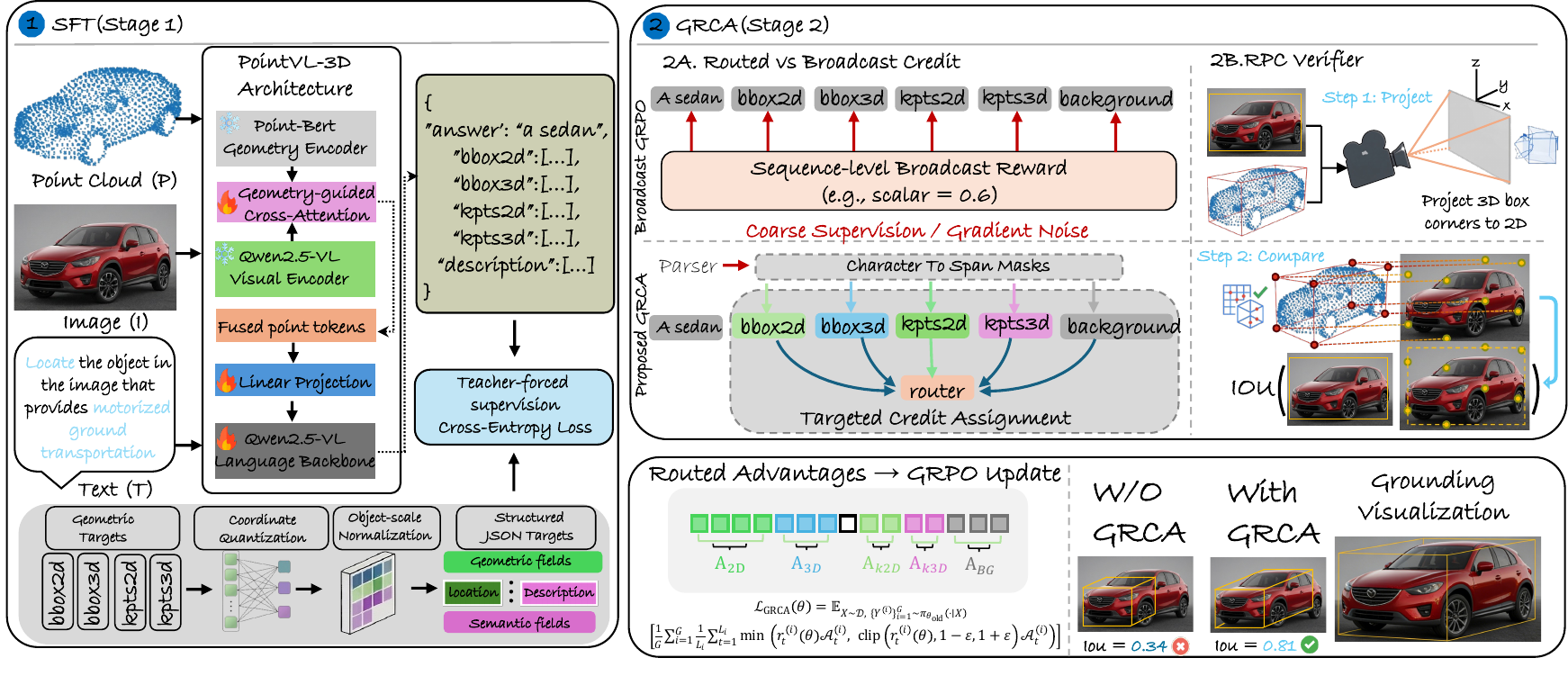}
\caption{Overview of PointVL-3D. In Stage 1, a hybrid point-image architecture fuses Point-BERT geometry features with the Qwen2.5-VL backbone and is supervised to generate structured JSON outputs. In Stage 2, GRCA replaces sequence-level broadcast reward with field-specific routed credit assignment over geometric token spans, while an RPC verifier enforces global consistency by comparing the predicted 2D box with the reprojection of the predicted 3D box.}
\Description{A two-stage framework for PointVL-3D showing supervised fine-tuning for structured JSON generation and GRCA post-training with routed field-specific rewards and reprojection consistency.}
\label{fig:framework}
\end{figure*}

\subsection{Benchmarks for Grounded Spatial Reasoning}
Grounded spatial reasoning in 3D has been studied through both question answering and referring expression benchmarks. ScanQA~\cite{azuma2022scanqa} and SQA3D~\cite{ma2023sqa3d} combine scene understanding with language reasoning, while ScanRefer~\cite{chen2020scanrefer} and ReferIt3D~\cite{achlioptas2020referit3d} evaluate language-guided localization in real 3D scenes. These benchmarks establish the standard task formats for grounded perception and spatial reasoning.

Recent datasets broaden this landscape toward more realistic grounding regimes. Multi3DRefer~\cite{zhang2023multi3drefer} extends grounding beyond the single-target setting and requires predicting zero, one, or multiple valid referents. EmbodiedScan~\cite{wang2024embodiedscan} introduces ego-centric scene perception with language-grounded tasks for embodied agents. ScanReason~\cite{zhu2024scanreason} further shifts the task from direct description matching to reasoning-driven grounding under implicit instructions. 3D-GRAND and its accompanying 3D-POPE benchmark~\cite{yang20253d} emphasize grounding quality and hallucination in 3D-LLMs at larger scale. These benchmarks make progress on task diversity, but most existing protocols still score the final answer and the localized target in separate spaces. As a result, a model can perform reasonably in 2D or 3D when measured independently while still failing to produce one spatial hypothesis that is jointly compatible with the image evidence and the recovered 3D geometry. Our evaluation targets exactly this missing layer of supervision by testing whether the predicted 3D structure remains valid after calibrated reprojection. The goal is not only to measure whether the answer is close to the annotation, but also whether it can stand as a geometrically coherent explanation across views and modalities.

\subsection{Post-Training and Credit Assignment for Structured Outputs}
Post-training methods such as RLHF~\cite{ouyang2022training}, DPO~\cite{rafailov2023direct}, and GRPO~\cite{shao2024deepseekmath} are now standard tools for aligning generative models with downstream objectives. Their common assumption is that a training signal can be assigned to the generated sequence at a relatively coarse level. This is often sufficient when quality is judged by overall preference or semantic adequacy.

Structured geometric prediction makes this assumption less suitable. In our setting, only a small subset of tokens determines whether the final prediction is spatially correct, while the remaining tokens mainly carry instruction context, natural language scaffolding, or output formatting. Recent work on finer-grained supervision, including process reward models~\cite{guo2025segment} and token-level value estimation~\cite{tran2025exploiting}, seeks to improve credit assignment granularity, but typically relies on additional decomposition, auxiliary estimation, or denser forms of supervision. Our setting admits a more direct solution because the output schema already reveals which fields are responsible for geometric validity. This allows us to assign reward to the exact token spans that encode the spatial prediction rather than spreading the same signal across the full sequence. The benefit is not only better efficiency. More importantly, it aligns optimization with the actual error surface of the task, where a small coordinate mistake should matter more than a large number of semantically harmless language tokens.

\subsection{Geometric Consistency and Verifier-Guided 3D Grounding}
Geometric consistency is a classical principle in multi-view geometry, reconstruction, and SLAM, where reprojection agreement provides a physically grounded test of correctness. In learning-based systems, related ideas are often used indirectly through fusion, filtering, or consistency checks over multi-view observations. ConceptFusion~\cite{jatavallabhula2023conceptfusion} is a representative example that uses multi-view fusion to build open-set multimodal 3D maps, showing the value of geometry-aware aggregation for downstream reasoning.

Recent 3D grounding methods also make increasing use of explicit view selection and geometric evidence at inference time. LLM-Grounder~\cite{yang2024llm} uses an LLM to decompose complex grounding queries and reason over candidates proposed by external 3D grounding tools. SeeGround~\cite{li2025seeground} converts 3D scenes into query-aligned rendered views and spatially enriched descriptions so that strong 2D vision-language models can be applied to zero-shot 3D grounding. These methods show that better use of geometry and viewpoint can substantially improve grounding, especially in open-vocabulary settings. However, geometry in these systems mainly acts after candidate generation as a mechanism for selection, ranking, or validation. Our work moves this signal into training itself. Instead of using geometric consistency only to filter or refine outputs after they are produced, we use it to shape the policy toward predictions that are already more likely to satisfy the scene geometry before any external correction is applied. In that sense, the verifier is not only an inference-time checker but part of the learning objective.

\section{Method}
\label{sec:method}

We study structured spatial generation for point-aware vision-language models. Given an input
\(
X=(P,I,T)
\),
where \(P\) denotes a point cloud, \(I\) an RGB image, and \(T\) a text instruction, the model generates an autoregressive output sequence
\(
Y=(y_1,\ldots,y_L)\sim \pi_\theta(\cdot \mid X)
\)
that follows a fixed JSON schema. The output contains four geometric fields,
\begin{equation}
\mathcal{F}
=
\{\texttt{bbox2d},\texttt{bbox3d},\texttt{kpts2d},\texttt{kpts3d}\},
\label{eq:fields}
\end{equation}
together with two semantic text fields, \texttt{answer} and \texttt{description}. Our goal is to improve geometric alignment during post-training by replacing sequence-level reward broadcasting with routed, field-specific credit assignment.

\subsection{Hybrid Point-Image Policy}
\label{sec:arch}

We construct a hybrid policy by combining a 3D geometry encoder with a 2D vision-language backbone. Let
\(
H_P \in \mathbb{R}^{M\times d}
\)
be the point tokens extracted by Point-BERT~\cite{yu2022point}, and let
\(
H_I \in \mathbb{R}^{K\times d}
\)
be the image tokens produced by the visual encoder of Qwen2.5-VL~\cite{bai2025qwen2}. To inject texture and appearance cues into the point representation, we apply geometry-guided cross-attention:
\begin{equation}
H_{\mathrm{cross}}
=
\mathrm{Softmax}\!\left(
\frac{(H_PW_Q)(H_IW_K)^\top}{\sqrt d}
\right)(H_IW_V).
\label{eq:cross_attn_new}
\end{equation}
The fused representation is formed as
\begin{equation}
\hat H_P = H_P + H_{\mathrm{cross}},
\label{eq:fused_tokens}
\end{equation}
and is then projected into the token space of the language backbone by a learnable linear map,
\begin{equation}
Z_P = \hat H_P W_{\mathrm{proj}},
\label{eq:proj_tokens}
\end{equation}
where \(W_{\mathrm{proj}} \in \mathbb{R}^{d \times d_{\mathrm{LM}}}\). The resulting visual tokens \(Z_P\) are concatenated with the instruction \(T\), and the whole sequence is decoded autoregressively by the policy \(\pi_\theta\).

\subsection{Stage 1: Structural Warm-up}
\label{sec:sft}

Before reinforcement learning, the model is first trained to obey the target schema and to produce coarse but executable geometric outputs. We enforce a fixed JSON template containing two semantic fields (\texttt{answer}, \texttt{description}) and four geometric fields (\texttt{bbox2d}, \texttt{bbox3d}, \texttt{kpts2d}, \texttt{kpts3d}). We quantize continuous coordinates into discrete bins \(b \in \{0,1,\ldots,1000\}\) and serialize them into the output sequence. Given a ground-truth sequence \(Y^*=(y_1^*,\ldots,y_{L^*}^*)\), we optimize the standard supervised objective
\begin{equation}
\mathcal{L}_{\mathrm{SFT}}(\theta)
=
-
\mathbb{E}_{(X,Y^*)\sim\mathcal D}
\left[
\sum_{t=1}^{L^*}
\log \pi_\theta(y_t^* \mid y_{<t}^*, X)
\right].
\label{eq:sft_new}
\end{equation}
This warm-up stage teaches the model the output format and a coarse input-output correspondence, but it does not explicitly distinguish structurally important coordinate tokens from ordinary language tokens.

\subsection{Stage 2: Geometric Reward Credit Assignment}
\label{sec:grca}

The core difficulty in post-training is that geometric correctness depends on a small subset of coordinate tokens, whereas conventional RL methods typically broadcast a single scalar reward over the entire sequence. We address this mismatch with Geometric Reward Credit Assignment (GRCA), which parses the generated structure, computes field-specific rewards, and routes each reward only to the token span responsible for it.

\paragraph{Parsing and span masks.}
For a decoded sample \(Y\), we parse the JSON structure and recover the character spans of each geometric field. These character spans are then mapped back to token indices using a robust character-to-token alignment procedure based on cumulative decoded offsets. For each parsed sample \(Y\), this yields a disjoint partition
\begin{equation}
\{\mathcal I_f(Y)\}_{f\in\mathcal F}
\quad\text{and}\quad
\mathcal I_{\mathrm{bg}}(Y),
\label{eq:partition_new}
\end{equation}
where \(\mathcal I_f(Y)\) is the token span of geometric field \(f\), and \(\mathcal I_{\mathrm{bg}}(Y)\) contains all remaining tokens, including semantic fields such as \texttt{answer} and \texttt{description}, as well as formatting and structural tokens. In our routing scheme, all tokens in \(\mathcal I_{\mathrm{bg}}(Y)\) receive only the background advantage.
\begin{equation}
m_{t,f}(Y)=\mathbf 1[t\in \mathcal I_f(Y)],
\qquad
m_{t,\mathrm{bg}}(Y)=\mathbf 1[t\in \mathcal I_{\mathrm{bg}}(Y)].
\label{eq:masks}
\end{equation}

\paragraph{Field-specific rewards.}
Before reward computation, the generated coordinate tokens are parsed from the JSON output and de-quantized from discrete bins back to continuous geometric values. Each geometric field is evaluated by a task-specific reward
\(
r_f(Y)\in[0,1]
\).
For bounding boxes, we use Intersection over Union (IoU). For keypoints, we use a containment-based reward. Let \(\{\hat{k}_m\}_{m=1}^M\) denote the predicted keypoints, and let \(B_f^*\) denote the corresponding ground-truth bounding box for the keypoint field \(f \in \{\texttt{kpts2d}, \texttt{kpts3d}\}\). We define
\begin{equation}
r_{\mathrm{kpt}}(Y)
=
\frac{1}{M}\sum_{m=1}^{M}
\mathbf{1}\!\left(\hat{k}_m \in B_f^*\right),
\label{eq:kpt_reward}
\end{equation}
where \(M\) is the number of predicted keypoints. For \texttt{kpts2d}, \(B_f^*\) denotes the ground-truth 2D bounding box and the containment test is performed on the image plane. For \texttt{kpts3d}, \(B_f^*\) denotes the ground-truth 3D bounding box and the containment test is performed in 3D space. This reward measures whether the predicted keypoints fall inside the target box rather than whether they match ground-truth keypoints one by one. In evaluation, we report Keypoint Accuracy (KPA) under the same criterion in 2D and 3D. These rewards remain field-local, and each reward supervises only the token span that generated the corresponding field.

\paragraph{Group-relative standardization.}
Following GRPO~\cite{shao2024deepseekmath}, for each input \(X\) we sample a group of \(G\) candidate outputs
\(
\{Y^{(i)}\}_{i=1}^G \sim \pi_{\theta_{\mathrm{old}}}(\cdot\mid X)
\).
For field \(f\), let
\(
\mu_f
\)
and
\(
\sigma_f
\)
denote the mean and standard deviation of \(\{r_f(Y^{(i)})\}_{i=1}^G\). We compute the standardized field advantage as
\begin{equation}
\tilde A_f^{(i)}
=
\frac{r_f(Y^{(i)})-\mu_f}{\sigma_f+\epsilon}.
\label{eq:field_adv_new}
\end{equation}
This avoids training a separate value network and makes heterogeneous reward scales comparable across fields.

\paragraph{Routed token-wise advantages.}
Instead of assigning one global advantage to all tokens, we construct a routed token-wise advantage for each sampled output:
\begin{equation}
\mathcal A_t^{(i)}
=
\sum_{f\in\mathcal F}
m_{t,f}\!\left(Y^{(i)}\right)\tilde A_f^{(i)}
+
m_{t,\mathrm{bg}}\!\left(Y^{(i)}\right)\tilde A_{\mathrm{bg}}^{(i)}.
\label{eq:routed_adv}
\end{equation}
Thus, tokens belonging to \texttt{bbox3d} receive only the standardized reward of \texttt{bbox3d}, tokens belonging to \texttt{kpts3d} receive only the keypoint reward, and so on. This preserves field-relevant credit on geometric spans and reduces direct interference from unrelated reward components.

\subsection{Reprojection Consistency as a Global Verifier}
\label{sec:rpc}

Field-wise rewards improve local accuracy, but they do not by themselves guarantee that the predicted 3D geometry is consistent with the predicted 2D grounding. To encourage cross-modal consistency, we introduce a reprojection-consistency reward.

The predicted field \texttt{bbox3d} is decoded into an axis-aligned 3D bounding box \(B_{\mathrm{3D}}(Y)\). Let \(\{c_j\}_{j=1}^8\) denote its eight corners. Given camera intrinsics \(K\) and extrinsics \((R,t)\), each corner is projected onto the image plane as
\begin{equation}
\tilde{u}_j = K(Rc_j + t),
\qquad
u_j =
\left(
\frac{\tilde{u}_{j,1}}{\tilde{u}_{j,3}},
\frac{\tilde{u}_{j,2}}{\tilde{u}_{j,3}}
\right),
\label{eq:projection_new}
\end{equation}
where \(\tilde{u}_j\) is the homogeneous image coordinate and \(u_j \in \mathbb{R}^2\) is the corresponding 2D point.

We define the reprojection of \(B_{\mathrm{3D}}(Y)\) as the smallest axis-aligned 2D box enclosing all projected corners,
\begin{equation}
B_{\mathrm{proj}}(Y)
=
\Big[
\min_j u_{j,x},\;
\min_j u_{j,y},\;
\max_j u_{j,x},\;
\max_j u_{j,y}
\Big],
\label{eq:bproj_new}
\end{equation}
where \(u_{j,x}\) and \(u_{j,y}\) denote the horizontal and vertical coordinates of \(u_j\), respectively.

Let \(B_{\mathrm{2D}}(Y)\) denote the predicted 2D bounding box in the same output. We then define the reprojection-consistency reward as
\begin{equation}
r_{\mathrm{rpc}}(Y)
=
\mathrm{IoU}\!\left(B_{\mathrm{2D}}(Y),\, B_{\mathrm{proj}}(Y)\right).
\label{eq:rpc_new}
\end{equation}

We standardize \(r_{\mathrm{rpc}}\) within the group to obtain \(\tilde A_{\mathrm{rpc}}^{(i)}\). Since RPC depends jointly on multiple geometric fields and can be noisy near projection boundaries, we do not inject it directly into a single coordinate span. Instead, we use it to shape the background tokens:
\begin{equation}
\tilde A_{\mathrm{bg}}^{(i)}
=
(1-\lambda)\left(
\frac{1}{|\mathcal F|}
\sum_{f\in\mathcal F}\tilde A_f^{(i)}
\right)
+
\lambda\,\tilde A_{\mathrm{rpc}}^{(i)}.
\label{eq:bgadv_new}
\end{equation}
This design treats RPC as a global preference for prediction-level 2D--3D consistency, rather than as a local reward on any single coordinate field.

\subsection{Routed GRPO Objective}
\label{sec:objective}

Our training objective follows the clipped policy-ratio form used in GRPO, but replaces the broadcast sequence-level advantage with the routed token-wise advantage in Eq.~\eqref{eq:routed_adv}. For token \(t\) in sample \(Y^{(i)}\), we define the policy ratio as
\begin{equation}
r_t^{(i)}(\theta)
=
\frac{
\pi_\theta\!\left(y_t^{(i)} \mid y_{<t}^{(i)}, X\right)
}{
\pi_{\theta_{\mathrm{old}}}\!\left(y_t^{(i)} \mid y_{<t}^{(i)}, X\right)
}.
\label{eq:ratio_new}
\end{equation}

The routed surrogate objective is
\begin{equation}
\begin{aligned}
\mathcal L_{\mathrm{GRCA}}(\theta)
&=
\mathbb E_{
X \sim \mathcal D,\;
\{Y^{(i)}\}_{i=1}^G \sim \pi_{\theta_{\mathrm{old}}}(\cdot\mid X)
}
\Bigg[
\frac{1}{G}\sum_{i=1}^G
\frac{1}{L_i}\sum_{t=1}^{L_i}
\min\!\Big(
r_t^{(i)}(\theta)\mathcal A_t^{(i)}, \\
&\qquad\qquad\qquad
\mathrm{clip}(r_t^{(i)}(\theta),1-\varepsilon,1+\varepsilon)\mathcal A_t^{(i)}
\Big)
\Bigg],
\end{aligned}
\label{eq:final_obj_new}
\end{equation}
where \(L_i\) is the generated sequence length of sample \(Y^{(i)}\), and \(\varepsilon\) is the clipping threshold. In implementation, we maximize this surrogate objective, or equivalently minimize its negative.

Relative to vanilla GRPO, the optimization form remains unchanged, but the advantage is no longer broadcast uniformly across the full sequence. Instead, each token receives the routed advantage associated with its corresponding field or background role. This makes the credit assignment better matched to the structured output. Because the objective averages over tokens, longer fields contribute proportionally more updates; in our setting, we retain this behavior since fields with more coordinates naturally carry more geometric information.

\subsection{Analysis of Routed Credit Assignment}
\label{sec:theory}

We give a simple conditional moment analysis to illustrate how routing affects the direct gradient contribution on a field-specific span. The purpose is not to prove global convergence, but to compare the direct gradient contribution on a field-specific token span under broadcast versus routed credit assignment. Unless otherwise stated, all conditional moments in this subsection are taken with respect to the sampling distribution induced by the rollout policy for a fixed input \(X\).

For a parsed sample \(Y\), define the score restricted to field \(f\) as
\begin{equation}
S_f(Y)
=
\sum_{t\in \mathcal I_f(Y)}
\nabla_\theta \log \pi_\theta(y_t \mid y_{<t},X).
\label{eq:scoref_new}
\end{equation}
Let \(\bar A(Y)\) denote a sequence-level broadcast advantage, and let \(\tilde A_f(Y)\) be the field-specific routed advantage. Write
\begin{equation}
\bar A(Y)=\tilde A_f(Y)+\xi_f(Y),
\label{eq:decomp_new}
\end{equation}
where \(\xi_f(Y)\) collects the residual contribution from other fields and background preferences. The direct gradient term on span \(\mathcal I_f(Y)\) is then
\begin{align}
\hat G_f^{\mathrm{broad}}
&=
\bar A(Y)S_f(Y), \nonumber\\
\hat G_f^{\mathrm{route}}
&=
\tilde A_f(Y)S_f(Y).
\label{eq:gbroad_groute_new}
\end{align}
Therefore,
\begin{equation}
\hat G_f^{\mathrm{broad}}
=
\hat G_f^{\mathrm{route}}+\xi_f(Y)S_f(Y).
\label{eq:broad_route_relation_new}
\end{equation}

If
\begin{equation}
\mathbb E[\xi_f(Y)S_f(Y)\mid X]=0,
\label{eq:orthogonality_new}
\end{equation}
then broadcast and routed training have the same conditional mean direct contribution on field \(f\):
\begin{equation}
\mathbb E[\hat G_f^{\mathrm{broad}}\mid X]
=
\mathbb E[\hat G_f^{\mathrm{route}}\mid X].
\label{eq:mean_compare_new}
\end{equation}

To compare fluctuation magnitude for vector-valued gradients, we define
\begin{equation}
\mathrm{Var}[Z]
:=
\mathrm{tr}(\mathrm{Cov}[Z]),
\qquad
\mathrm{SCov}[U,V]
:=
\mathrm{tr}(\mathrm{Cov}[U,V]).
\label{eq:var_def_new}
\end{equation}
If
\begin{equation}
\mathrm{SCov}\!\left(
\tilde A_f(Y)S_f(Y),\,
\xi_f(Y)S_f(Y)\mid X
\right)\ge 0,
\label{eq:cov_assump_new}
\end{equation}
then
\begin{equation}
\mathrm{Var}\!\left[\hat G_f^{\mathrm{broad}}\mid X\right]
\ge
\mathrm{Var}\!\left[\hat G_f^{\mathrm{route}}\mid X\right].
\label{eq:var_compare_new}
\end{equation}
Moreover, using Eq.~\eqref{eq:broad_route_relation_new},
\begin{align}
\mathrm{Var}\!\left[\hat G_f^{\mathrm{broad}}\mid X\right]
&=
\mathrm{Var}\!\left[\hat G_f^{\mathrm{route}}\mid X\right]
+
\mathrm{Var}\!\left[\xi_f(Y)S_f(Y)\mid X\right]
\nonumber\\
&\quad+
2\,\mathrm{SCov}\!\left(
\tilde A_f(Y)S_f(Y),\,
\xi_f(Y)S_f(Y)\mid X
\right).
\label{eq:var_decomp_new}
\end{align}
Thus, for the direct contribution on a field-specific span, broadcast credit assignment carries an additional residual term from unrelated fields, whereas routed credit assignment removes that term by construction. This comparison is limited to the direct contribution on \(\mathcal I_f(Y)\). It does not imply full decoupling, since autoregressive dependencies and shared parameters can still couple different spans indirectly.

\subsection{Robustness and Edge Cases}
\label{sec:robustness}

The routing mechanism depends on successful parsing. If a field is missing, duplicated, malformed, or unparsable, we set its span to the empty set and assign zero reward to that field. This discourages invalid outputs without introducing ambiguous supervision. The overall training pipeline is initialized from pre-trained Point-BERT and Qwen2.5-VL weights, followed by the SFT stage and then the routed GRPO stage.

\section{Experiments and Results}
\label{sec:exp}

\subsection{Dataset Construction and Protocols}
We evaluate on a synthetic, camera-calibrated benchmark derived from ShapeNetCore~\cite{chang2015shapenet}. Unlike standard datasets that rely on low-fidelity raw renders, we construct a Generative-Refined 3D-VL Benchmark containing 12,347 high-quality instances across 51 object categories. The construction pipeline comprises three stages:

\subsubsection{Point Cloud and Geometry Processing}
We perform data cleaning to merge semantically overlapping classes and remove categories with insufficient generation quality. To convert meshes into high-quality point clouds $P \in \mathbb{R}^{N \times 3}$, we employ Poisson Disk Sampling ($N{=}1024$, init factor 5) via Open3D. This ensures a uniform distribution of points on the manifold surface, preserving geometric topology better than random sampling. For models with missing normals, we perform KNN-based estimation ($k{=}30$).

\subsubsection{Generative Visual Synthesis Pipeline}
To bridge the domain gap between synthetic renders and real-world images, we implement a tri-stage visual synthesis pipeline. First, we use PyTorch3D for adaptive rasterization. We design a Category-Adaptive View Sampling strategy where elevation is constrained to $[15^\circ, 35^\circ]$ and azimuth is tailored to object semantics (e.g., side-view for cars) to prevent meaningless viewpoints. Second, we employ Qwen-Image-Edit \cite{wu2025qwenimagetechnicalreport} as a semantic low-pass filter. By prompting for anti-aliased and photorealistic surfaces, we reconstruct discrete jagged edges into continuous curves while maintaining geometric alignment. Finally, we generate photorealistic images using FLUX.1-Krea-dev \cite{flux1kreadev2025} controlled by ControlNet-Union-Pro \cite{zhang2023adding}. We implement a Control Guidance Termination strategy where spatial constraints are applied only during the first 80\% of diffusion steps to balance structural fidelity with textural naturalness. Strict manual verification was performed on tri-stage visualizations to remove samples with geometric distortion or pose drift, resulting in a final split of 11,347 training and 1,000 testing samples.

\begin{table}[t]
\centering
\scriptsize
\setlength{\tabcolsep}{3.2pt}
\renewcommand{\arraystretch}{1.10}
\rowcolors{3}{NatRow}{white}

\begin{tabular}{@{}p{0.56\columnwidth}cc@{}}
\toprule
\rowcolor{NatHeader}
Model &
\cellcolor{NatBlue1}IoU$_{2D}$ $\uparrow$ &
\cellcolor{NatPurple}KPA-2D $\uparrow$ \\
\midrule
Qwen2.5-VL-3B \cite{bai2025qwen2} & 0.791 & 0.657 \\
Qwen2.5-VL-7B \cite{bai2025qwen2}  & 0.780 & 0.751 \\
InternVL3\_5-4B-Instruct \cite{wang2025internvl3} & 0.772 & 0.727 \\
InternVL3\_5-8B-Instruct \cite{wang2025internvl3} & 0.764 & 0.680 \\
DeepSeek-VL2-Tiny \cite{wu2024deepseek} & 0.692 & 0.182 \\
MiMo-VL-7B-RL \cite{li2025xiaomi} & 0.745 & 0.601 \\
Qwen2.5-VL-3B(SFT) & 0.891 & 0.882 \\
MiMo-VL-7B-RL(SFT) & 0.886 & 0.887 \\
\rowcolor{NatOurs}
PointVL-3D (Ours) & 0.894 & 0.890 \\
\bottomrule
\end{tabular}
\caption{Comparison with open-source 2D vision-language baselines on the test split. General-purpose VLMs are evaluated only on 2D grounding metrics because they do not produce native 3D structured outputs.}
\label{tab:vlm_2d}
\end{table}

\begin{table}[t]
\centering
\scriptsize
\setlength{\tabcolsep}{3.2pt}
\renewcommand{\arraystretch}{1.10}
\rowcolors{3}{NatRow}{white}

\begin{tabular}{@{}p{0.56\columnwidth}ccc@{}}
\toprule
\rowcolor{NatHeader}
Model &
\cellcolor{NatGreen1}IoU$_{3D}$ $\uparrow$ &
\cellcolor{NatCyan}KPA-3D $\uparrow$ &
\cellcolor{NatPeach}RPC $\uparrow$ \\
\midrule
PointNet \cite{qi2017pointnet} & 0.492 & 0.923 & 0.810 \\
PointNet++ \cite{qi2017pointnet++} & 0.456 & 0.884 & 0.820 \\
PointBert \cite{yu2022point} & 0.440 & 0.856 & 0.780 \\
Point Transformer V3 \cite{wu2024point} & 0.238 & 0.885 & 0.521 \\
Point-Language-3D & 0.542 & 0.567 & 0.814 \\
\rowcolor{NatOurs}
PointVL-3D (Ours) & 0.686 & 0.930 & 0.852 \\
\bottomrule
\end{tabular}
\caption{Comparison with representative 3D-capable baselines on structured spatial grounding over the test split.}
\label{tab:vlm_3d}
\end{table}

\subsubsection{Automated Multi-Modal Annotation}
We establish a scalable annotation engine to generate dense supervision. We utilize GPT-5 to generate fine-grained attribute QA pairs and geometric descriptions. For 2D grounding, we deploy Qwen2.5-VL-30B-Instruct with temperature 0.0 to localize the most prominent instance. For 3D grounding, we compute axis-aligned bounding boxes (AABB) and employ Farthest Point Sampling (FPS) to select 16 topological keypoints. 2D keypoints are generated via morphological foreground mask processing. These annotated keypoints are used to construct the structured supervision targets in Stage 1. In post-training and evaluation, the keypoint reward and KPA follow the containment criterion defined in Sec.~\ref{sec:grca}.

\subsection{Implementation Details}
\label{sec:imp_details}
Our model is built upon the pre-trained Qwen2.5-VL-3B-Instruct \cite{bai2025qwen2} backbone. The geometry encoder uses Point-BERT pre-trained on ShapeNet, freezing the original layers and training only the cross-attention layers. The input point cloud is downsampled to $N=1024$ points, and images are resized to $512 \times 512$.

We implement our framework using PyTorch and DeepSpeed Zero-3 \cite{rasley2020deepspeed}. In Stage 1 (SFT), we fine-tune for 2 epochs with a batch size of 128, using the AdamW optimizer with a cosine learning rate schedule ($2e^{-5}$ to $2e^{-7}$). In Stage 2 (GRCA), we perform RL post-training for 1 epoch with a group size $G=8$ and a learning rate of $1e^{-5}$. The RPC mixing weight is set to $\lambda=0.15$. All experiments are conducted on 4 NVIDIA A800 (80GB) GPUs, taking approximately 48 hours.
\begin{table}[t]
\centering
\scriptsize
\setlength{\tabcolsep}{2.8pt}
\renewcommand{\arraystretch}{1.10}
\rowcolors{3}{NatRow}{white}

\begin{tabularx}{\linewidth}{Xccccc}
\toprule
\rowcolor{NatHeader}
Method &
\cellcolor{NatBlue1}IoU$_{2D}$ $\uparrow$ &
\cellcolor{NatGreen1}IoU$_{3D}$ $\uparrow$ &
\cellcolor{NatPurple}KPA-2D $\uparrow$ &
\cellcolor{NatCyan}KPA-3D $\uparrow$ &
\cellcolor{NatPeach}RPC $\uparrow$ \\
\midrule
SFT & 0.890 & 0.663 & 0.890 & 0.640 & 0.836 \\
SFT (budget-matched) & 0.888 & 0.678 & 0.880 & 0.670 & 0.844 \\
GRPO (broadcast reward) & 0.886 & 0.684 & 0.890 & 0.890 & 0.841 \\
GRCA & 0.894 & 0.683 & 0.890 & 0.930 & 0.846 \\
\rowcolor{NatHi}
GRCA + RPC term & 0.894 & 0.686 & 0.890 & 0.930 & 0.852 \\
\bottomrule
\end{tabularx}
\caption{Main quantitative results under a fixed post-training budget, comparing supervised fine-tuning, broadcast-reward GRPO, and the proposed GRCA variants.}
\label{tab:main}
\end{table}

\subsection{Tasks and Evaluation Metrics}
We evaluate the model along two primary tasks and one consistency metric.

Structured Spatial Grounding: For geometric evaluation, we focus on four predicted fields: \texttt{bbox2d}, \texttt{bbox3d}, \texttt{kpts2d}, and \texttt{kpts3d}. Performance is quantified by Intersection over Union (IoU) for bounding boxes and by Keypoint Accuracy (KPA) for keypoints. For both 2D and 3D keypoints, KPA is defined as the fraction of predicted keypoints that fall inside the corresponding ground-truth bounding box.

Visual Question Answering (VQA): We evaluate multiple-choice category prediction and attribute-based QA accuracy to assess semantic understanding.

Cross-Modal Consistency Metric: We report Reprojection Consistency (RPC), computed as the 2D IoU between the predicted 2D box and the projected footprint of the predicted 3D box. A high RPC score indicates that the model's 3D spatial reasoning is causally aligned with its 2D visual perception.

\subsection{Comparative Analysis}

\subsubsection{Performance vs. 2D VLMs}
We first benchmark our approach against leading open-source VLMs on 2D grounding tasks in Table~\ref{tab:vlm_2d}. General-purpose models like Qwen2.5-VL-3B achieve respectable zero-shot performance (0.791 IoU$_{2D}$) but remain weaker under our structured 2D evaluation protocol, scoring only 0.657 on KPA-2D. Our PointVL-3D significantly outperforms this baseline, improving KPA-2D to 0.890 and IoU$_{2D}$ to 0.894. Compared to the supervised fine-tuned baseline (Qwen2.5-VL-3B SFT), our method maintains comparable or slightly better 2D performance. This indicates that our RL-based alignment does not suffer from the "forgetting" phenomenon often observed in multi-objective optimization; instead, the geometric constraints likely reinforce the visual localization features.

\subsubsection{Performance vs. 3D-Capable Models}
Table~\ref{tab:vlm_3d} illustrates the fundamental limitations of existing 3D architectures. Pure point-cloud models like PointNet achieve strong KPA-3D under our evaluation criterion, suggesting that they capture object geometry reasonably well, but they still lag behind in 3D box localization. Conversely, multimodal baselines like Point-Language-3D leverage semantic priors to improve IoU$_{3D}$ (0.542) but suffer from modality misalignment, as reflected by their lower KPA-3D and RPC scores. Our PointVL-3D framework effectively unifies these strengths. By achieving 0.686 IoU$_{3D}$ and 0.930 KPA-3D, it surpasses the specialized PointNet in its own domain while maintaining high semantic accuracy. The high RPC score (0.852) serves as strong evidence that our model has learned to align 3D predictions with 2D visual evidence, minimizing geometric hallucinations.

\subsection{Ablation Studies}

\subsubsection{Mechanism of Improvement: Signal-to-Noise Ratio}
Table~\ref{tab:main} dissects the contribution of each component. SFT alone plateaus at 0.640 KPA-3D, suggesting that maximum likelihood estimation struggles to provide sufficiently direct supervision for keypoint placement. Standard GRPO (broadcast reward) improves this to 0.890 but is fundamentally limited by credit assignment ambiguity. By explicitly routing rewards via GRCA, we boost KPA-3D to 0.930. Although training uses a field-local geometric validity reward for keypoints, the resulting improvements consistently transfer to the evaluation-time KPA metric, indicating better keypoint placement under the containment criterion. The critical comparison is against the "SFT (budget-matched)" baseline, which only reaches 0.670. This large gap (0.260) confirms that the performance leap stems from the high signal-to-noise ratio of our targeted reinforcement signal, rather than merely extended computational budget. Finally, the RPC term acts as a verifier, lifting the consistency metric to 0.852.

\begin{table*}[t]
\centering
\small
\setlength{\tabcolsep}{3.2pt}
\renewcommand{\arraystretch}{1.12}
\rowcolors{4}{NatRow}{white}

\begin{tabular*}{\textwidth}{@{\extracolsep{\fill}}lccccccccc@{}}
\toprule
\rowcolor{NatHeader}
Model &
\multicolumn{4}{c}{\cellcolor{NatBlue2}BLEU $\uparrow$} &
\multicolumn{3}{c}{\cellcolor{NatGreen2}ROUGE $\uparrow$} &
\cellcolor{NatSand}BERTScore $\uparrow$ &
\cellcolor{NatPeach}QA Acc. $\uparrow$ \\
\cmidrule(lr){2-5}\cmidrule(lr){6-8}
\rowcolor{NatHeader}
-- & B-1 & B-2 & B-3 & B-4 & R-1 & R-2 & R-L & -- & -- \\
\midrule 
InternVL3\_5-4B-Instruct\cite{wang2025internvl3}
 &
-- & -- & -- & -- &
-- & -- & -- & 0.8230 & 0.850 \\
Qwen2.5-VL-3B\cite{bai2025qwen2}
 &
-- & -- & -- & -- &
-- & -- & -- & 0.8220 & 0.742 \\
Qwen2.5-VL-3B\cite{bai2025qwen2}
(SFT) & 0.5418	& 0.3902 & 0.2706 & 0.1799 & 0.5249 & 0.2143 & 0.4051 & 0.9147 & 0.951 \\

MiMo-VL-7B-RL\cite{li2025xiaomi}
 &
-- & -- & -- & -- &
-- & -- & -- & 0.8209 & 0.876 \\
MiMo-VL-7B-RL\cite{li2025xiaomi}
(SFT) & 0.5425 & 0.3933 & 0.2754 & 0.1837 & 0.5312 & 0.2148 & 0.4113 & 0.9162 & 0.953 \\
SFT &
0.5457 & 0.3917 & 0.2725 & 0.1819 &
0.5225 & 0.2171 & 0.4093 & 0.9154 & 0.953 \\
GRCA &
0.5436 & 0.3890 & 0.2685 & 0.1767 &
0.5300 & 0.2140 & 0.4075 & 0.9152 & 0.951 \\
\rowcolor{NatHi}
GRCA + RPC term &
0.5483 & 0.3948 & 0.2737 & 0.1823 &
0.5335 & 0.2183 & 0.4122 & 0.9154 & 0.951 \\
\bottomrule
\end{tabular*}
\caption{Text generation metrics and question-answering accuracy on the test split, used to assess whether geometric post-training preserves language performance. Entries marked by ``--'' indicate that the corresponding model does not yield valid outputs for this evaluation unless it is fine-tuned on the task. We therefore do not report these metrics.}
\label{tab:nlp}
\end{table*}

\begin{figure}[t]
\centering
\includegraphics[width=\columnwidth]{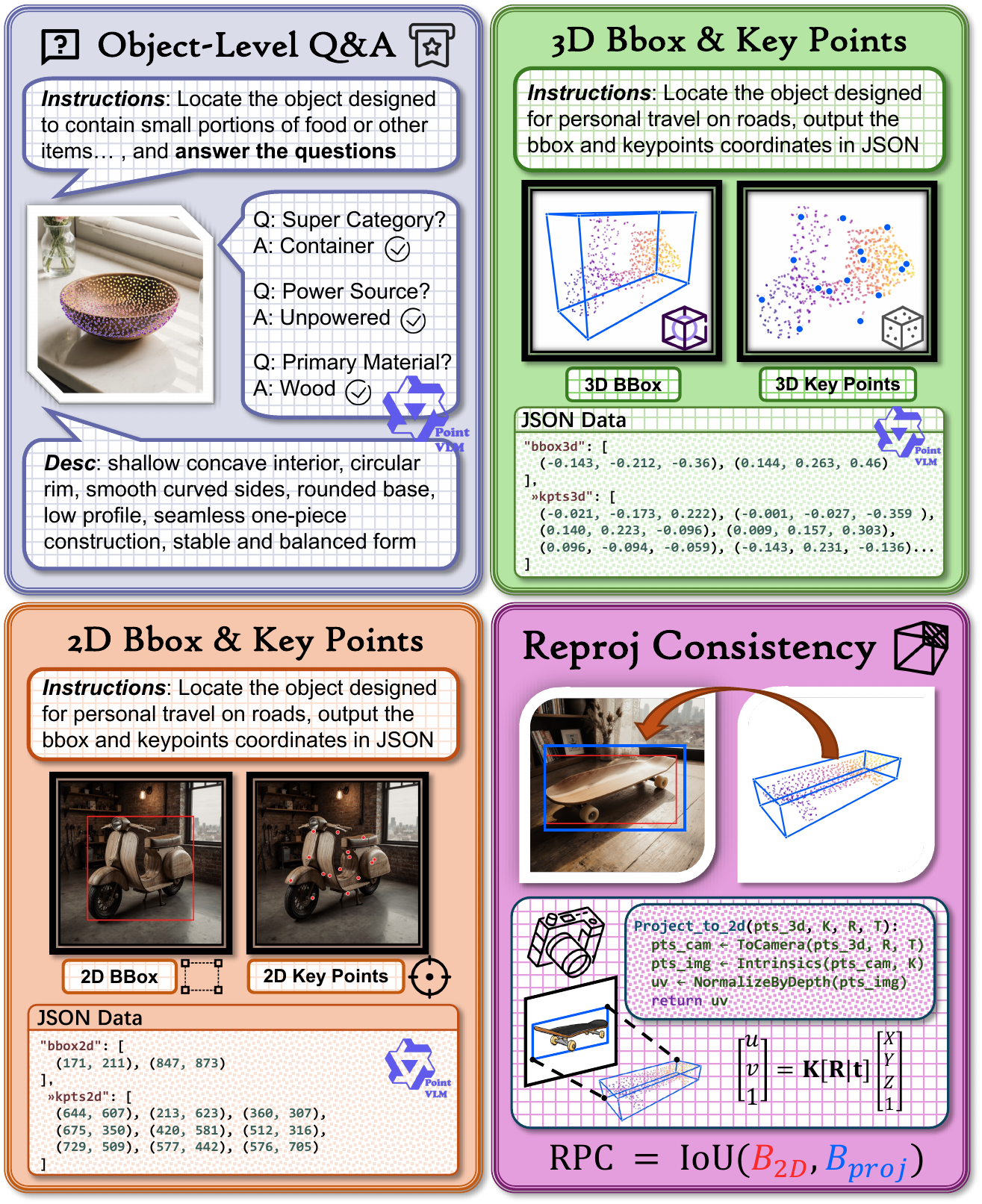}
\caption{Visualization of structured outputs in our framework. The figure shows examples of object-level question answering, 3D bounding box and keypoint prediction, 2D bounding box and keypoint prediction, and the reprojection consistency computation.}
\Description{A four-panel figure showing object-level question answering, 3D bounding box and keypoint prediction, 2D bounding box and keypoint prediction, and reprojection consistency computed by projecting 3D geometry into the image plane and comparing it with the predicted 2D box.}
\label{fig:visualization}
\end{figure}

\subsubsection{Sensitivity and Over-Regularization}
Table~\ref{tab:rpc_weight} analyzes the sensitivity to the consistency weight $\lambda$. We observe an optimal balance at $\lambda=0.15$, yielding the highest RPC (0.852). Notably, increasing $\lambda$ beyond 0.175 causes a drop in KPA-3D and IoU$_{2D}$ (e.g., KPA-3D drops to 0.910 at $\lambda=0.20$). We attribute this to over-regularization: when the consistency constraint becomes dominant, the policy may prioritize generating 3D boxes that project perfectly into the 2D mask but deviate from the true 3D geometry, effectively "hacking" the projection reward at the expense of metric accuracy.

\begin{table}[t]
\centering
\scriptsize
\setlength{\tabcolsep}{2.8pt}
\renewcommand{\arraystretch}{1.10}
\rowcolors{3}{NatRow}{white}

\begin{tabular}{@{}lccccc@{}}
\toprule
\rowcolor{NatHeader}
Setting &
\cellcolor{NatBlue1}IoU$_{2D}$ $\uparrow$ &
\cellcolor{NatGreen1}IoU$_{3D}$ $\uparrow$ &
\cellcolor{NatPurple}KPA-2D $\uparrow$ &
\cellcolor{NatCyan}KPA-3D $\uparrow$ &
\cellcolor{NatPeach}RPC $\uparrow$ \\
\midrule
GRCA (w/o RPC term)             & 0.894 & 0.683 & 0.890 & 0.930 & 0.846 \\
GRCA + RPC term, $\lambda$=0.10 & 0.894 & 0.684 & 0.890 & 0.930 & 0.839 \\
GRCA + RPC term, $\lambda$=0.125 & 0.893 & 0.683 & 0.890 & 0.930 & 0.847 \\
\rowcolor{NatHi}
GRCA + RPC term, $\lambda$=0.15 & 0.894 & 0.686 & 0.890 & 0.930 & 0.852 \\
GRCA + RPC term, $\lambda$=0.175 & 0.893 & 0.677 & 0.890 & 0.930 & 0.845 \\
GRCA + RPC term, $\lambda$=0.20 & 0.885 & 0.684 & 0.890 & 0.910 & 0.844 \\
\bottomrule
\end{tabular}
\caption{Sensitivity analysis of the RPC mixing weight $\lambda$ on grounding accuracy and reprojection consistency.}
\label{tab:rpc_weight}
\end{table}

\subsubsection{Zero Alignment Tax on Language}
A common concern in RLHF is the "alignment tax," where optimizing for a specific reward degrades general capabilities. Table~\ref{tab:nlp} shows that our method maintains statistical parity with the SFT baseline across BLEU, ROUGE, and BERTScore metrics. This stability is consistent with our GRCA design, which localizes field-specific reward signals to their corresponding geometric spans while treating semantic tokens mainly through the background preference. This reduces unnecessary interference with language generation and preserves linguistic fluency.

\subsection{Structured Output Visualization}
\label{sec:visualization}

Figure~\ref{fig:visualization} shows examples of the structured outputs in our framework. The upper-left panel presents object-level question answering from a functional instruction together with semantic attributes and a geometric description. The upper-right and lower-left panels show examples of the predicted 3D and 2D grounding outputs, including bounding boxes, keypoints, and their JSON representations. The lower-right panel illustrates the reprojection consistency computation by projecting predicted 3D geometry into the image plane and comparing it with the predicted 2D box using IoU.

\section{Conclusion}
\label{sec:conclusion}

In this work, we presented PointVL-3D, a unified framework designed to bridge the gap between semantic understanding and geometric executability. Identifying the impedance mismatch between sparse geometric tokens and dense reinforcement signals, we proposed Geometric Reward Credit Assignment (GRCA). This mechanism routes precise metric feedback directly to the corresponding token spans, effectively transforming generic policy optimization into targeted structural alignment. Furthermore, we integrated a Reprojection-Consistency (RPC) verifier, enabling the model to internalize geometric constraints and reduce geometric hallucinations.

Extensive experiments on our Generative-Refined benchmark demonstrate strong performance gains, boosting 3D KPA from 0.64 to 0.93 while maintaining 2D localization capabilities. These results suggest that with precise credit assignment, Vision-Language Models can evolve from passive describers into verifiable spatial reasoners.

Currently, the reliance on explicit camera parameters and synthetic data distributions limits immediate in-the-wild deployment. Future work will focus on learning joint camera estimation to relax calibration requirements, bridging the sim-to-real gap, and extending GRCA to dynamic robotic manipulation tasks.

\balance
\bibliographystyle{ACM-Reference-Format}
\bibliography{references}

\end{document}